# A Lightweight Multi-Cancer Tumor Localization Framework for Deployable Digital Pathology

Brian Isett, PhD[1,2,*]; Rebekah Dadey, PhD [1,2]; Aofei Li, MD[3,+]; Ryan C. Augustin, MD[1,2,++]; Kate Smith, MS[1]; Aatur D. Singhi, MD[3]; Qiangqiang Gu, PhD[3,*]; Riyue Bao, PhD[1,2,*]

[1]UPMC Hillman Cancer Center, Pittsburgh, PA, USA; [2]Malignant Hematology and Medical Oncology Division, Department of Medicine, University of Pittsburgh, Pittsburgh, PA, USA; [3]Department of Pathology, University of Pittsburgh, Pittsburgh, PA, USA; [+]Current address: Division of Dermatopathology, Indiana University, Indianapolis, IN, USA; [++]Current address: Hematology and Medical Oncology Division, Mayo Clinic, Rochester, MN, USA;
[*]Corresponding authors: B.I., bri8@pitt.edu; Q.G., qug14@pitt.edu; R.B., rib37@pitt.edu

**Abstract**
*Accurate localization of tumor regions from hematoxylin and eosin-stained whole-slide images is fundamental for translational research including spatial analysis, molecular profiling, and tissue architecture investigation. However, deep learning-based tumor detection trained within specific cancers may exhibit reduced robustness when applied across different tumor types. We investigated whether balanced training across cancers at modest scale can achieve high performance and generalize to unseen tumor types. A multi-cancer tumor localization model (MuCTaL) was trained on 79,984 non-overlapping tiles from four cancers (melanoma, hepatocellular carcinoma, colorectal cancer, and non-small cell lung cancer) using transfer learning with DenseNet169. The model achieved a tile-level ROC-AUC of 0.97 in validation data from the four training cancers, and 0.71 on an independent pancreatic ductal adenocarcinoma cohort. A scalable inference workflow was built to generate spatial tumor probability heatmaps compatible with existing digital pathology tools. Code and models are publicly available at https://github.com/AivaraX-AI/MuCTaL.*

**Introduction**
The adoption of whole-slide imaging (WSI) in digital pathology workflows has enabled large-scale computational analysis of tumor morphology and microenvironment architecture from routine hematoxylin and eosin (H&E) histology slides[1,2]. Manual tumor annotation on WSIs is labor-intensive and often infeasible at scale, motivating the development of automated tumor localization methods[3–7]. Deep learning approaches have substantially advanced tumor detection in histopathology, with many studies demonstrating high performance in specific cancer types. For example, in stomach cancer and colorectal cancer (CRC), convolutional neural networks (CNNs) were used to identify tumor and predict clinically relevant molecular phenotypes such as microsatellite instability directly from H&E slides, supporting the concept that tumor-associated morphological patterns can be learned from routine histology images[8]. In skin cancer, CNNs achieved dermatopathologist-level competency in classifying melanoma (MEL) and other malignant lesions from benign tissue[9], and collective intelligence of machine-learning algorithms outperformed experts on the diagnosis task of pigmented lesions[10]. In hepatocellular carcinoma (HCC), the inception V3 network[11] demonstrated high accuracy on predicting malignancy, tumor differentiation, and mutated genes[12]. Similarly, in non-small cell lung cancer (NSCLC), deep learning models trained on H&E WSIs have shown automated identification of regions containing neoplastic cells and subsequent slide-level disease subtype classification[13,14] and survival prediction[15,16].

A persistent challenge in computational pathology is the limited generalizability of models trained on single-cancer datasets. Variability in tissue morphology, staining protocols, slide preparation, and scanner characteristics can introduce substantial domain shift between datasets, leading to degraded performance when models are applied to slides originating from different tumor types[17,18]. Conversely, large-scale foundation models trained on thousands of WSIs have demonstrated impressive model generalizability across cancers[19–26]. While these models represent a major step toward universal histopathologic feature extraction, their development typically requires extensive multi-institutional data aggregation, centralized harmonization, and substantial computational infrastructure that may not be readily available in many translational research settings[1,27].

Recent studies have shown that deep learning models trained on histopathology images from one cancer type can capture morphological features that generalize across other tumor types without requiring foundation-scale datasets, revealing conserved spatial patterns in tumor histology[28]. In practice, research teams at community hospitals and academic medical centers often work with limited, heterogeneous datasets assembled across multiple independent projects over time[29]. Developing computational frameworks that leverage balanced, modest-scale multi-cancer training cohorts therefore represents a practical and scalable approach for translational applications[29], yet remains underexplored.

Here, we investigate whether a lightweight multi-cancer training strategy can provide sufficient morphological diversity to support robust tumor localization across heterogeneous histopathology datasets, while remaining computationally tractable. We developed a multi-cancer tumor localization (MuCTaL) model trained on image tiles derived from four tumor types (MEL, HCC, CRC, and NSCLC), and evaluated its performance within each cancer as well as its ability to generalize to an unseen tumor type, pancreatic ductal adenocarcinoma (PDAC). To support integration with digital pathology workflows, we generated heatmap-based visualization to highlight tumor regions within WSIs and export the spatial coordinates of tile-wise classes compatible with open-source tools such as QuPath[30], providing interpretable insights into tumor localization within complex histologic landscapes.

**Materials and Methods**

*Study cohorts.* Human specimens of MEL, HCC, and PDAC were collected at University of Pittsburgh Medical Center (UPMC). Specimens were evaluated by two pathologists (A.L. for MEL, A.S. for HCC and PDAC) and one oncologist (R.C.A.). Published datasets were obtained for CRC[31] and NSCLC[32]. All patients were consented prior to sample collection. The study was approved by the University of Pittsburgh Institutional Review Board (IRB) (IRB18-177).

*H&E staining and image acquisition.* Data were generated for the institutional cohorts following published protocols[33–35]. All staining procedures were performed at the UPMC Translational Oncologic Pathology Services (TOPS). In brief, formalin-fixed paraffin-embedded (FFPE) tissue blocks were sectioned at 4 μm thickness, and slides were baked at 60 °C for one hour prior to processing. Sections were subsequently cooled, deparaffinized, and rehydrated in distilled water. H&E staining was performed using Hematoxylin 560 MX (Cat# 3801576) and Eosin Phloxine 515 (Cat# 3801606) with Define MX-aq (Cat# 3803598) and Blue Buffer 8 (Cat# 3802918) reagents (Leica Biosystems), according to the manufacturer's protocols. Stained slides were digitized at 40× magnification using a Leica AT2 whole-slide scanner to generate high-resolution digital images for this study.

*Training, validation, and test sets.* We created a dataset of 79,984 tiles from the four tumor types, with 90% for training, and 10% reserved for validation. An independent cohort of PDAC (7,346 tiles) was used as test set.

*Training dataset construction.* We constructed a multi-cancer tile-level training dataset across four tumor types (MEL, HCC, CRC, NSCLC). From each cancer, non-overlapping image tiles (224x224 pixels) were extracted from annotated tumor and non-tumor regions on WSIs. Preprocessing included several steps: (1) tiles were filtered to remove artifacts and blank ones, retaining only tiles with >70% tissue, (2) blurred tiles and tiles containing blood clots were removed using OpenCV[36]-based image quality filtering, and (3) color normalization and stain augmentation were applied using the Macenko method[37]. To minimize class imbalance and ensure comparable representation across tumor types, we resampled tiles to generate a balanced dataset with a 50:50 tumor /non-tumor tile ratio with approximately 20,000 tiles per tumor type, totaling 79,984 tiles across four tumor types for training. Tile extraction and preprocessing were implemented using the PathML framework[38] (v2.1.1) for standardized slide-to-tile conversion. To ensure compatibility between published and institutional cohorts, evaluation was performed at the tile level, which is a commonly used approach when involving publicly available histopathology tile datasets. For the institutional cohorts where de-identified patient IDs were available (MEL and HCC), data was balanced by patient within each tumor type, ensuring equal representation of patients among the ~20k tiles per cancer via resampling.

*Model architecture and training procedure.* We implemented a CNN using transfer learning from a pretrained DenseNet169 backbone in PyTorch[39]. The output layer was modified to perform binary classification (tumor versus non-tumor). The model was trained for 10 epochs with early layer weights frozen, followed by 20 epochs of fine tuning with max learning rate (LR) of $3e^{-3}$, 5 epochs with base LR $1e^{-5}$ and 5 epochs with base LR of $5e^{-5}$ to estimate learning rate gradient adjustments. During training, tiles were pre-sized, randomly rotated up to +/- 45 degrees, flipped and cropped with no re-scaling as part of data augmentation. Training was performed on an Nvidia A100 GPU with "batch_size = 375" using the Fastai (v2.7) deep learning framework[40]. All training jobs were executed on the high-performance computing (HPC) clusters at the University of Pittsburgh Center for Research Computing and Data (CRCD).

*Inference procedure.* To enable scalable deployment on WSIs, we built a distributed inference workflow by assigning each WSI to a separate job within the HPC environment using a SLRUM scheduler at the University of Pittsburgh CRCD. For each slide, non-overlapping tiles were extracted using the same preprocessing pipeline applied during training. Tile-level tumor detection probabilities were computed using the trained classifier. Predicted probabilities were spatially reassembled to generate slide-level tumor probability heatmaps. To improve spatial coherence, probability maps were smoothed using Gaussian filtering prior to thresholding. Tumor regions were delineated by applying a probability threshold of 0.5 (reflecting the 50:50 class distribution during training), followed by extraction of contiguous regions using contour detection. Identified tumor contours were rescaled to the original slide resolution and exported as GeoJSON objects compatible with QuPath[30] (v5.0).

*Model evaluation strategy.* Model performance was evaluated at the tile level on the validation and test sets using F1 score, sensitivity, specificity, and receiver operating characteristic-area under the curve (ROC-AUC). Performance metrics were computed both overall and stratified by tumor type to assess variability across datasets. Tile-level classification metrics were computed to quantify performance.

**Results**

The overall workflow for multi-cancer tumor localization is illustrated in **Fig. 1**. WSIs were partitioned into tiles, filtered for quality, normalized, and classified using DenseNet169 trained on a multi-cancer dataset with transfer learning (**Fig. 1A**). Tile-level tumor predictions were subsequently aggregated to generate spatial heatmaps of tumor detection probability across entire slides (**Fig. 1B**).

Across all validation tiles, the multi-cancer classifier achieved high overall performance with a ROC-AUC of 0.97, F1-score of 0.90, sensitivity of 0.94, and specificity of 0.86 (**Fig. 2A**). These results indicate that the model can accurately distinguish tumor from non-tumor regions at the tile level across heterogeneous cancer datasets.

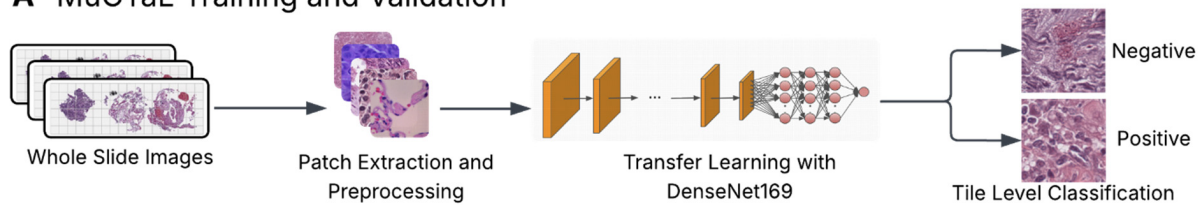
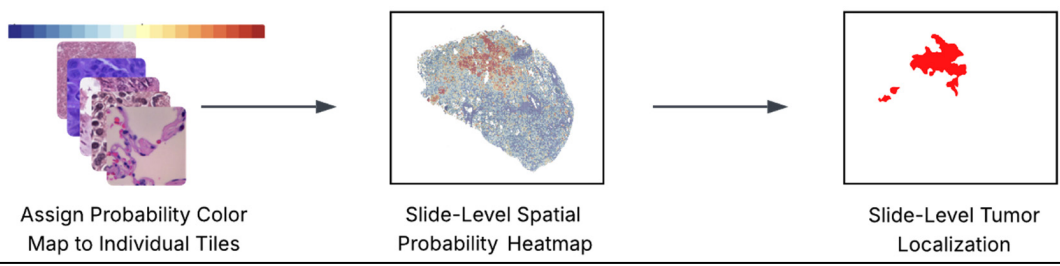

**Figure 1. MuCTaL framework for multi-cancer tumor localization.** (**A**) Training and validation workflow. Whole-slide images (WSIs) from multiple tumor types were partitioned into image tiles and subjected to preprocessing including tissue filtering, artifact removal, and stain normalization. Tiles were then used to train a convolutional neural network (CNN) using transfer learning from a pretrained DenseNet169 architecture to classify tumor versus non-tumor at tile level. (**B**) Slide-level visualization workflow. Tile-level tumor probabilities were assigned color values and spatially reconstructed to generate spatial heatmaps and contiguous tumor localization masks across the whole slide.

When stratified by tumor type, classification performance varied across datasets (**Fig. 2B; Table 1**). The model achieved near-perfect performance for CRC (AUC = 0.9999) and NSCLC (AUC = 1.00), with corresponding F1 scores approaching 1.0. MEL tiles also demonstrated high performance (AUC = 0.96, F1 = 0.89). Performance was lower for HCC (AUC = 0.79, F1 = 0.74). Examination of misclassification rates revealed that the highest proportion of misclassified tiles occurred in the HCC cohort (27%), followed by MEL (11%), whereas CRC and NSCLC cohorts exhibited minimal errors. To evaluate whether the multi-cancer classifier shows generalization, the model was applied to an independent dataset of PDAC slides that were not included in training, achieving an AUC of 0.71 (**Fig. 2B, Table 1**). Together, these findings demonstrated that multi-cancer training accurately detected tumor localization and retained the ability to identify malignant morphology that extend beyond individual tumor types.

Beyond classification, we reconstructed spatial tumor localization across WSIs. As shown in **Fig. 2C**, tile-level tumor probabilities were aggregated to generate slide-level heatmaps. These heatmaps highlight tumor-enriched regions and spatial patterns of tumor localization within the surrounding tissue architecture. To facilitate downstream analysis, the probability maps were subsequently processed using Gaussian smoothing and thresholding to generate contiguous tumor masks. Extracted tumor contours were then rescaled to the original slide resolution and exported as GeoJSON objects. These results can be directly imported into open-source platforms such as QuPath for interactive exploration, providing a practical tool for integrative analysis with translational research workflows.

**Discussion**
In this study, we evaluated whether multi-cancer training at modest scale can support robust tumor localization in translational research environments. Our results demonstrate that introducing morphological diversity through multi-cancer training captures shared features of malignancy across tumor types while remaining computationally efficient. This strategy provides a practical framework for developing deployable tumor localization models that can operate across heterogeneous histopathology datasets.

Prior studies often focused on training individual cancer paradigms, using highly curated datasets such as Camelyon16 and Camelyon17[41–43], and the Breast Cancer Histology[44–46] (BACH) challenge datasets. While these models can achieve high performance within domain, their performance may degrade when applied to slides from different tumor types. These observations are consistent with broader findings that emphasizes domain shift remains a major challenge for generalizable histopathology models[17]. At the other end of the spectrum, large-scale models trained on tens of thousands of WSIs across diverse cancers have shown strong cross-cancer transferability. For example, a pan-cancer classifier trained on 27,000+ WSIs from 19 tumor types achieved a ROC-AUC of 0.99 on tumor/normal status[28]. Moreover, while foundation models offer powerful general-purpose representations[19–26], their deployment often relies on access to pretrained models and additional fine-tuning in specialized infrastructure, which may not always be readily available in translational research environments.

Our work occupies an intermediate space between these paradigms. Rather than maximizing scale, we investigated whether balanced sampling across multiple tumor types can support model robustness while maintaining a modest dataset size. In this context, multi-cancer training can be viewed as a data-efficiency strategy for learning tumor-associated histological patterns that potentially generalize across cancers. Consistent with this hypothesis, the multi-cancer classifier demonstrated high tumor localization performance across MEL, HCC, CRC, NSCLC and retained the ability to detect malignant morphology in PDAC.

We observed differential error rates across tumor cohorts, with increased misclassification in the institutional HCC cohort. This may reflect the inherent morphological heterogeneity of HCC tumors[47]. Similar challenges have been reported in prior studies, which highlight the sensitivity of histopathology models to tissue heterogeneity[18,48]. Proposed mitigation strategies include stain normalization, color augmentation, and domain-adversarial training[18,49]. While our current framework incorporated standard augmentation approaches, future work will explore explicit domain adaptation strategies to further improve cross-tumor generalization.

Beyond classification accuracy, we built a deployable inference workflow with the reconstruction of whole-slide tumor probability maps and tumor contours as GeoJSON objects compatible with digital pathology platforms. Our framework enables rapid tumor region identification and extraction without manual slide annotation for downstream spatial and molecular analyses, and facilitates the integration of tumor localization models into practical research workflows.

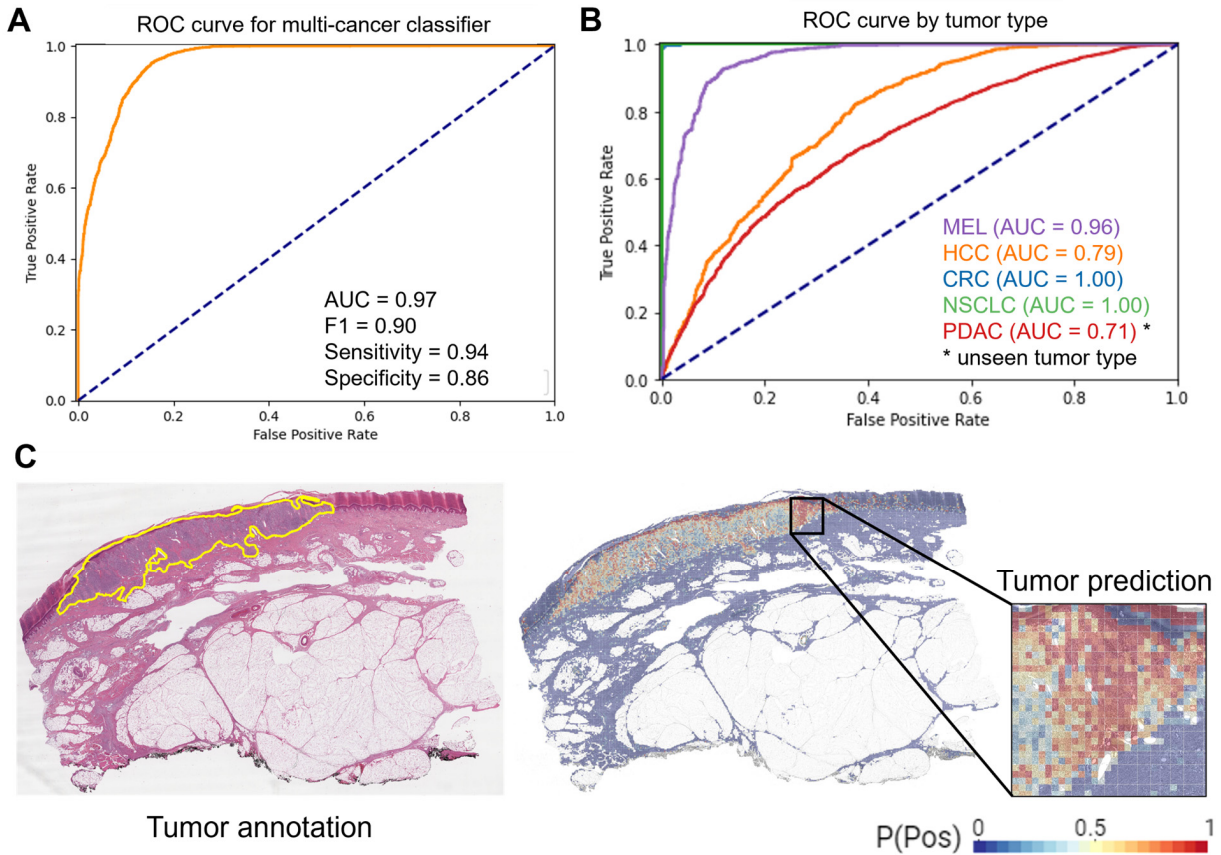

**Figure 2. Performance and spatial visualization of the multi-cancer tumor localization model.** (**A**) Receiver operating characteristic (ROC) curve for tile-level classification across all validation datasets. (**B**) ROC curves stratified by tumor type for melanoma (MEL), hepatocellular carcinoma (HCC), colorectal cancer (CRC), and non-small cell lung cancer (NSCLC). Pancreatic ductal adenocarcinoma (PDAC), which was not included during model training, was used as an independent evaluation dataset to assess model generalization. (**C**) Example of slide-level tumor localization generated from tile-level predictions (shown is MEL). Tumor probability scores were assigned to individual tiles and spatially reconstructed into a whole-slide probability heatmap. For the visualization of tumor-enriched regions within the tissue section, warmer colors indicate higher predicted probability of tumor presence. P(Pos) denotes predicted probability of tiles on the positive class (tumor).

**Table 1. Tile-level classification performance of the multi-cancer tumor localization classifier across tumor types.** Performance metrics are reported for each tumor cohort in the validation and test sets.

| Tumor Type | AUC | F1 Score | Sensitivity | Specificity | nTiles |
|---|---|---|---|---|---|
| MEL (melanoma) | 0.959 | 0.892 | 0.950 | 0.829 | 1,947 |
| HCC (hepatocellular carcinoma) | 0.786 | 0.744 | 0.813 | 0.628 | 2,000 |
| CRC (colorectal cancer) | 0.999 | 0.995 | 0.995 | 0.995 | 2,000 |
| NSCLC (non-small cell lung cancer) | 1.000 | 1.000 | 1.000 | 1.000 | 1,922 |
| PDAC (pancreatic ductal adenocarcinoma)* | 0.711 | 0.609 | 0.543 | 0.757 | 7,346 |

*PDAC represents an unseen tumor type that was not included during model training.

Several limitations should be noted. Our evaluation was conducted on five datasets and did not include large-scale multi-institution benchmarking. Slide-level metadata were not available for some public datasets used in this study, preventing strict enforcement of slide-level independence between training and validation tiles. Finally, although the model demonstrated generalization to an unseen tumor type, broader validation across additional cancers will be necessary to fully characterize cross-tumor robustness. Future studies will extend this framework to larger and more diverse datasets and systematically evaluate domain-shift effects across institutions and imaging platforms.

In conclusion, multi-cancer training at modest scale provides a practical strategy for tumor localization across heterogeneous cancer datasets. Lightweight multi-cancer models offer a feasible alternative to both single-cancer models and foundation-scale approaches, providing a scalable pathway for broader adoption of artificial intelligence-driven tumor localization in translational research environments.

### Acknowledgement
We thank the patients and families for their participation in this study. We thank Dr. Fangping Mu for technical assistance at the University of Pittsburgh Center for Research Computing and Data (CRCD) high-performance computing clusters (HPC). We acknowledge Hillman Career Acceleration Fellow for Innovative Cancer Research (R.B.) from the Hillman Program made possible by the Henry L. Hillman Foundation. **Funding.** This work was supported in part by National Cancer Institute (NCI) through the UPMC Hillman Cancer Center CCSG award P30CA047904 (R.B.), and The University of Pittsburgh CRCD through the resources provided, specifically the GPU clusters supported by NIH S10OD028483. This project used the UPMC HCC Cancer Bioinformatics Facility (CBS) and Translational Oncologic Pathology Services (TOPS). **Role of funding sources.** The funding sources had no role in the study design, data collection, data analysis, interpretation, or writing of the manuscript.

### Author's contributions
R.B. and B.I. conceived the study. R.B. supervised the study and provided funding. B.I. designed the methodology, implemented and built the codebase, and generated results, including image processing, model training, and validation. R.D. obtained specimens and organized data generation. A.L. performed pathology annotation in melanoma samples. A.D.S. performed pathology annotation in liver and pancreatic samples. R.C.A. provided oncology insights and reviewed clinical data. K.S. scanned the H&E slides and acquired the images. Q.G. organized and cleaned the codebase. B.I., Q.G., and R.B. wrote the manuscript. R.B. and Q.G. edited the manuscript. All authors provided feedback and approved the manuscript.

### Declaration of Competing Interests
R.B. declares PCT/US15/612657 (Cancer Immunotherapy), PCT/US18/36052 (Microbiome Biomarkers for Anti-PD-1/PD-L1 Responsiveness: Diagnostic, Prognostic and Therapeutic Uses Thereof), PCT/US63/055227 (Methods and Compositions for Treating Autoimmune and Allergic Disorders). Other authors declare that they have no known competing financial interests or personal relationships that could have appeared to influence the work reported in this paper.

### Data availability
Public datasets used in this study, including CRC and NSCLC datasets, are available from the original publications cited in the manuscript. Institutional datasets used in this study are not publicly available due to patient privacy and institutional data use restrictions. Correspondence and requests for materials should be addressed to R.B. (rib37@pitt.edu). Code and trained models for the MuCTaL framework are publicly available at https://github.com/AivaraX-AI/MuCTaL.